\documentclass[letterpaper, 10 pt, conference]{ieeeconf}  

\IEEEoverridecommandlockouts                              

\usepackage{graphicx} 
\usepackage{epsfig} 
\usepackage{mathptmx} 
\usepackage{times} 
\usepackage{amsmath} 
\usepackage{amssymb}  

\usepackage[font=small]{caption}
\usepackage{subcaption}

\usepackage{multirow}
\usepackage[usenames, dvipsnames]{color}

\usepackage{amsfonts}       
\usepackage{bm}
\usepackage{dsfont}
\usepackage{mathtools}
\usepackage{booktabs}
\usepackage{hyperref}
\usepackage[symbol]{footmisc}

\title{\LARGE \bf
Inferring Distributions Over Depth from a Single Image
}

\author{Gengshan Yang$^{1}$, Peiyun Hu$^{1}$, Deva Ramanan$^{1,2}$
\thanks{$^{1}$Robotics Institute, Carnegie Mellon University\qquad $^{2}$Argo AI
        }%
}

\begin{document}

\maketitle
\thispagestyle{empty}
\pagestyle{empty}

\begin{abstract}

When building a geometric scene understanding system for autonomous vehicles, it is crucial to know when the system might fail. Most contemporary approaches cast the problem as depth regression, whose output is a depth value for each pixel. Such approaches cannot diagnose when failures might occur. One attractive alternative is a deep Bayesian network, which captures uncertainty in both model parameters and ambiguous sensor measurements. However, estimating uncertainties is often slow and the distributions are often limited to be uni-modal. In this paper, we recast the continuous problem of depth regression as discrete binary classification, whose output is an un-normalized distribution over possible depths for each pixel. Such output allows one to reliably and efficiently capture multi-modal depth distributions in ambiguous cases, such as depth discontinuities 
and reflective surfaces.
Results on standard benchmarks show that our method produces accurate depth predictions and significantly better uncertainty estimations than prior art 
while running near real-time. Finally, by making use of uncertainties of the predicted distribution, we significantly reduce streak-like artifacts and improves accuracy as well as memory efficiency in 3D map reconstruction. Video and code can be found on the project website\footnote[1]{\href{https://github.com/gengshan-y/monodepth-uncertainty}{https://github.com/gengshan-y/monodepth-uncertainty}}.
\end{abstract}

\section{Introduction}


Most contemporary architectures for geometric scene understanding cast the problem as one of regression - given an image, infer a depth for each pixel. However, in safety-critical systems such as autonomous vehicles, such perceptual inferences will be used to make critical decisions and motion plans with considerable implications for safety. For example, what if the estimated depth of an obstacle on the road is incorrect? Here, it is crucial to build recognition systems that  (1) allow for safety-critical graceful-degradation in functionality, rather than catastrophic failures; (2) are self-aware enough to diagnose when such failures occur; and (3) extract enough information to take an appropriate action, e.g. a slow-down, pull-over, or alerting of a manual operator. Such requirements are explicitly laid out in Automotive Safety Integrity Level (ASIL) standards which self-driving vehicles will be required to satisfy~\cite{koopman2016challenges}.

\begin{figure}[t]
    \centering
    \includegraphics[width=1\linewidth]{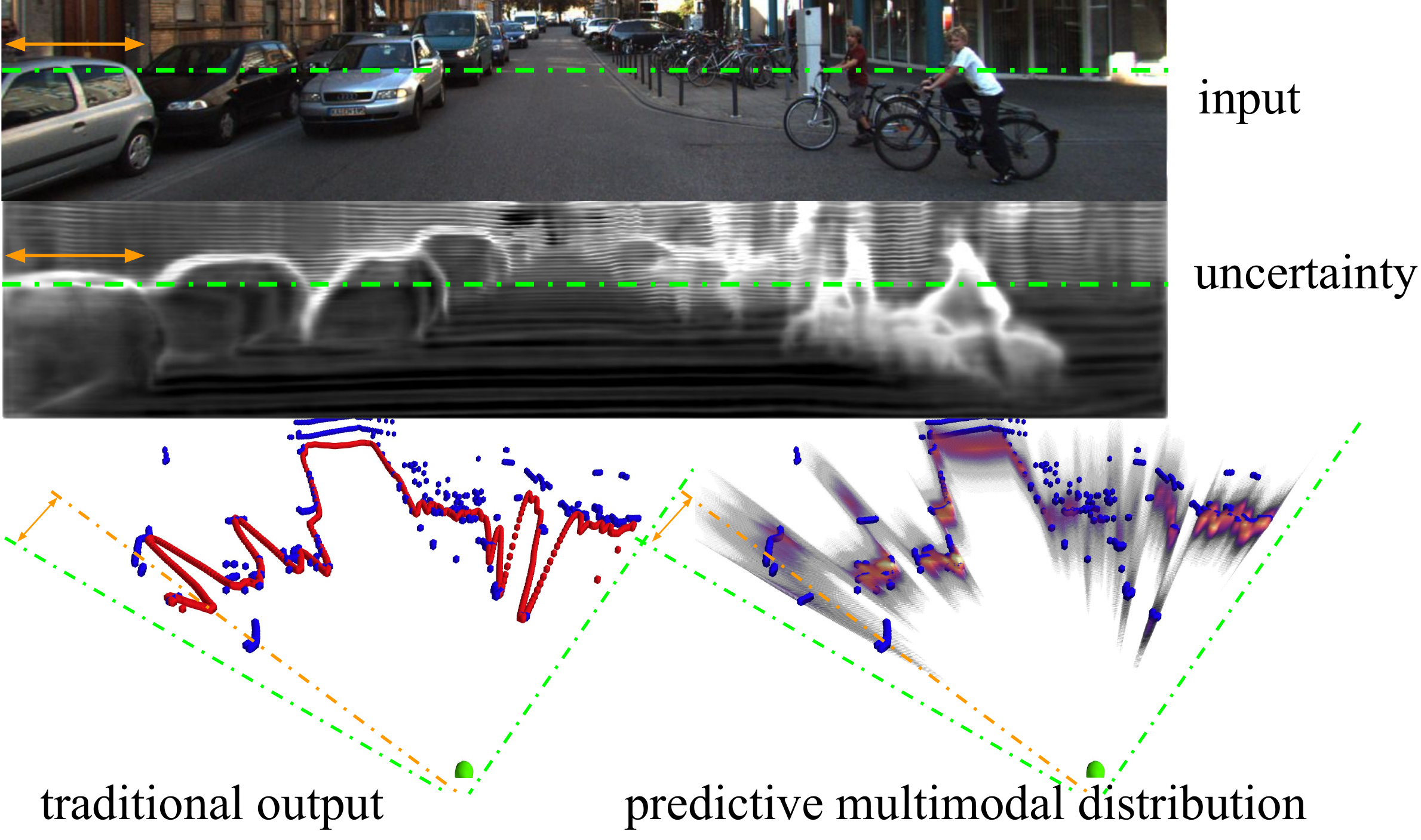}
    \setlength{\belowcaptionskip}{-14pt}
    \caption{Given an input image, traditional methods predict a single depth value for each pixel. In this paper, we describe an approach that predicts a per-pixel multi-modal distribution over depth. In the example above, we zoom in onto depth predictions along the dashed green line. Inside the input image, we highlight a segment filled with depth continuities marked with a yellow double-head arrow, where pixels could come from the car in the front, the car behind, or even the building in the back. In the output at the bottom, we mark ground truth depth with \textcolor{Blue}{blue} and depth with higher probabilities with \textcolor{Red}{red}. While traditional methods incorrectly yield the mean of different modes, our approach successfully captures the multi-modal nature. }
    \label{fig:firstfigure}
\end{figure}

\begin{figure*}
\centering
\includegraphics[width=\linewidth]{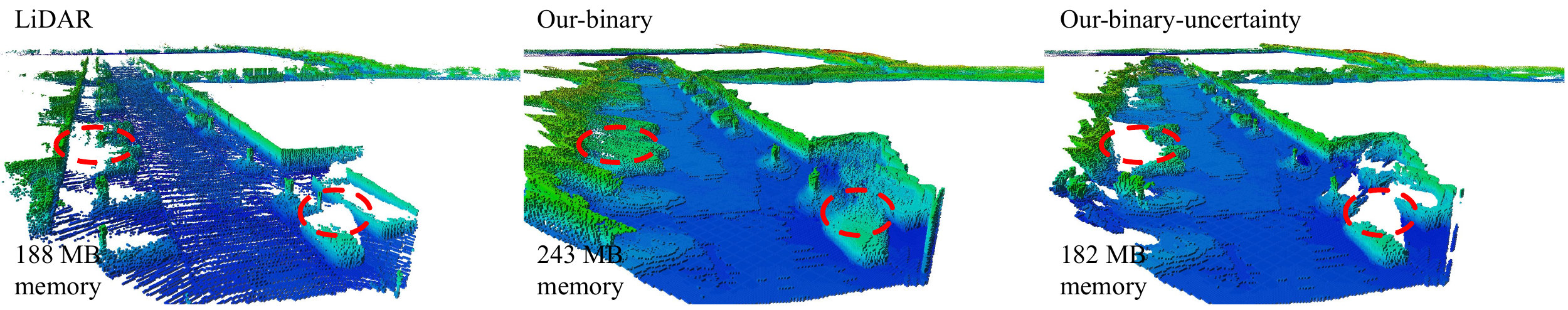}
\setlength{\belowcaptionskip}{-14pt}
\caption{From left to right, we show 3D maps built with LiDAR measurements (left), vanilla monocular depth predictions (middle), and {\it most certain} monocular depth predictions (right). Color encodes normalized heights.
  By thresholding depth predictions with uncertainty, we can remove streak-like artifacts (red dotted circles) and reduce memory usage by a quarter. To generate these maps, we feed depth measurements/predictions into OctoMap~\cite{hornung2013octomap} and use odometry measurements as provided. LiDAR and monocular images come from the KITTI odometry sequence-00, which is not included in training. }
\label{fig:mapping}
\end{figure*}

\begin{figure}
    \centering
     \includegraphics[width=\linewidth]{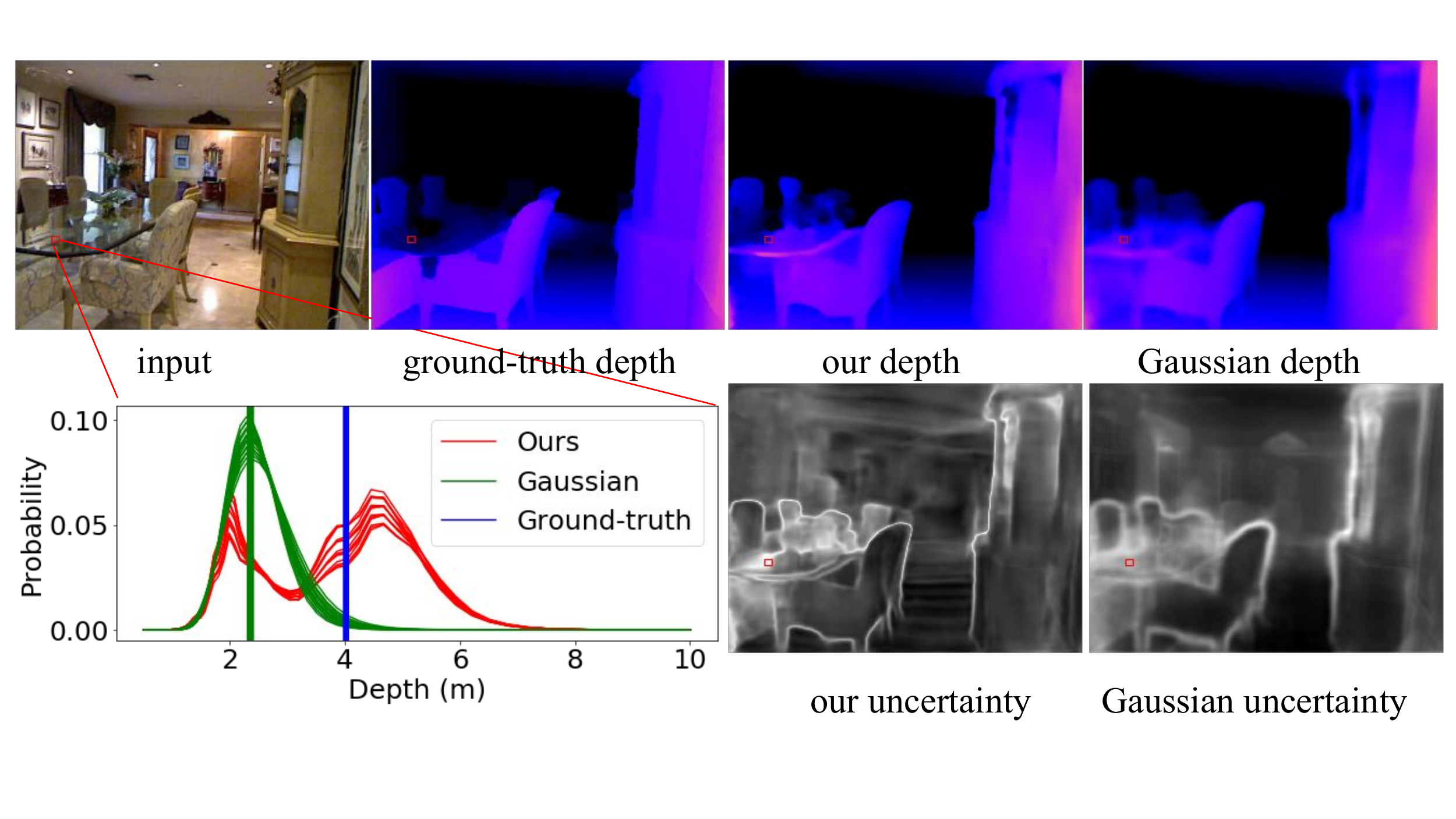}
    \caption{
    Visualization of multi-modal depth predictions on the glass table, where the surface is transparent, making its depth fundamentally ambiguous. Instead of regressing a single depth value or predicting a unimodal distribution, our method yields a multi-modal distribution over depth and successfully captures different modes (the table surface and the wall behind the table).}
        \label{fig:secondfigure}
       \vspace{-0.1in}
\end{figure}

Such safety standards represent significant challenges for data-driven machine vision algorithms, which are unlikely to provide formal guarantees of performance~\cite{DBLP:journals/corr/abs-1708-06374}. One attractive solution is that of probabilistic modeling, where uncertainty estimates are propagated throughout a model. In the contemporary world of deep learning, deep Bayesian methods~\cite{gal2016dropout,kendall2017uncertainties} provide uncertainty estimates over model parameters (e.g., observing a scene that looks different than experience) and uncertainty estimates arising from ambiguous data (e.g., a sensor failure). We apply such approaches to the problem of depth estimation from a single camera. Our particular approach differs from prior work in two notable aspects. First, prior methods often require Monte Carlo sampling to compute uncertainty estimates~\cite{gal2016dropout}, which can be slow for real-time safety-critical applications. Second, while certainty estimates provide some degree of self-awareness, they are limited to {\em uni}modal estimates of scene structure, implicitly producing a Gaussian estimate of depth represented by a regressed mean and regressed variance (or confidence)~\cite{kendall2017uncertainties}. Instead, we develop representations that report back {\em multi}modal distributions that allow us to ask more nuanced questions (e.g., ``what is the second possible depth of a pixel?'', ``how many modes exist in the distribution?''), as shown in Fig. \ref{fig:firstfigure} and Fig.~\ref{fig:secondfigure}.

\begin{figure*}
  \centering
  \includegraphics[trim=0 2 0 0, clip,width=1\linewidth]{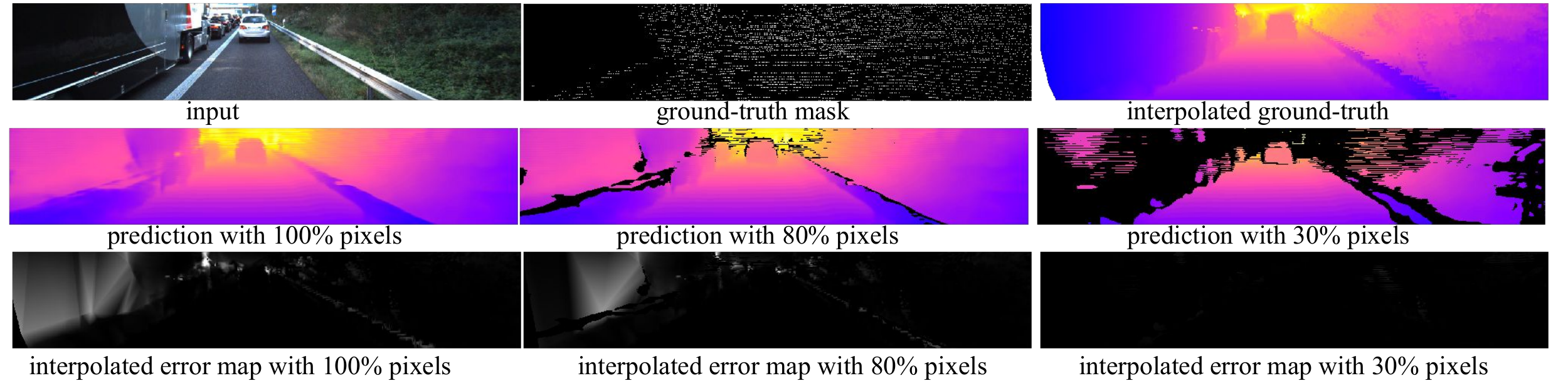}
 \setlength{\belowcaptionskip}{-14pt}
  \caption{A situation in which monocular depth estimation complements range sensing. In the top row, from left to right, we show a monocular image, a binary mask, and an entropy map. The binary mask shows where LiDAR readings are available and the entropy map summarizes the uncertainty of each pixel's predicted distribution.
    Note that a large chunk of the truck body with black paint has no LiDAR returns since LiDAR sensors are less reliable with less-reflective materials.
    Our monocular depth estimator successfully predicts high entropy in the area with black paint.
    In the bottom row, we show depth predictions with uncertain pixels removed. From left to right, we gradually increase the confidence threshold.
    The rightmost one plots 30\% pixels with the most confident depth predictions, in which we see most predictions on the truck body are removed.
    If a perception system solely relies on LiDAR measurements, it will perceive plenty of free space on the left side, which might lead to catastrophic decisions.
    If a perception system is designed with redundancy, it would trust LiDAR measurements less at pixels where the monocular estimator predicts uncertainties.
  }
  \label{fig:qualitative}

\end{figure*}

From a practical perspective, one may ask why bother estimating depth from a single camera when special-purpose sensors for depth estimation exist (such as LIDAR or multi-view camera rigs)? Common arguments include cost, payload and power consumption of robots~\cite{yang2017real}, but we motivate this problem from a safety perspective. One crucial method for ensuring ASIL certification is redundancy, and so estimates of scene geometry that are independently produced from various sensors (e.g., independently from LIDAR and independently from cameras) {\em and} that agree provide additional fault tolerance. In Fig.~\ref{fig:qualitative}, we illustrate a situation in which monocular depth estimation complements range sensing.

Our overall approach to probabilistic reasoning is to recast the continuous problem of depth regression (given an image patch $x$, regress a depth value $y \in R$) as a discrete problem of selecting one out of many possible discretized depths $y \in \{1,2, \ldots K\}$.  Previous work \cite{cao2017estimating} has already demonstrated that discretization can improve the {\em accuracy} of the underlying depth regression task, but we show that discretization is even more useful for producing simple and efficient (and possibility multimodal) {\em uncertainty} estimates of depth. Intuitively, K-way classifiers are often trained with softmax loss functions, and so naturally report a distribution over $K$ possible discrete depths.  Importantly, we find that such distributions can be further improved by recasting the multiclass formulation as a binary {\em multilabel} task  $y \in \{0,1\}^K$ -  essentially,  train $K$ independent binary classifiers that classify patches at particular discrete depths. It is straightforward to show that the binary multilabel formulation can be seen as a relaxation of the multiclass problem that removes a linear constraint. Removing this constraint creates a more challenging learning problem that appears to be better regularized in terms of uncertainty reports. At test-time, we use the $K$ logits as an unnormalized distribution over possible depths, though they can easily be normalized post-hoc (to compute summary statistics such as the expected depth).

Our main contributions are as follows:
\begin{itemize}
\item We formulate the problem of monocular depth estimation in a probabilistic framework, which gives us confidence intervals of depth instead of point estimations.
\item We recast the problem of depth regression as \textit{multi-label} depth classification, which yields \textit{reliable}, \textit{multi-modal} distributions over depth.
\item Our method produces accurate depth and significantly better uncertainty estimation over prior art on KITTI and NYU-depth while running {\it near real-time}.
\item Our predicted distribution over depths improves monocular 3D map reconstruction, reducing streak-like artifacts and improving accuracy as well as memory efficiency. 
\end{itemize}

\section{Related Work}

{\bf Single Image Depth Estimation:}
Early works~\cite{hoiem2005geometric,saxena2006learning} popularize the problem of inferring scene depth maps from a single image, making use of handcrafted features. Eigen et al.~\cite{eigen2014depth} take a data-driven approach to learn features in a coarse-to-fine network that refines global structure with local predictions. Some recent work substantially improves the performance of single image depth estimation using better deep neural network architectures \cite{laina2016deeper,liu2016learning,xu2017multi}.

{\bf Depth Estimation as Classification:}
Closely related to our work, Cao et al.~\cite{cao2017estimating} formulates depth estimation as a multi-class classification problem and use soft targets to train the model. However, they make inference by choosing the most likely depth class, which does not take full advantage of the depth distribution, while we explore richer inference methods based on the predicted depth distributions. More importantly, the standard multi-class classification approach tends to make confident errors and does not yield reliable uncertainty estimations. Instead, we learn the classification model as $K$ independent binary classifiers, which regularizes the model and gives us much better uncertainty estimation as well as noticeable performance improvement on standard benchmarks. Fu et al.~\cite{fu2018deep} formulate depth estimation as ordinal regression, aiming to predict a CDF over depth. However, they do not ensure the predicted CDF to be monotonically non-decreasing. This makes it ungrounded to apply probabilistic reasoning for uncertainty estimation. In contrast, we formulate depth estimation as a discrete classification problem, aiming to predict a valid depth PDF.

{\bf Uncertainty in Depth Estimation:}
Kendall et al.~\cite{kendall2017uncertainties} introduce two kinds of uncertainties: epistemic uncertainty (over model parameters) and aleatoric uncertainty (over output distributions). They show that epistemic uncertainty is data-dependent while aleatoric uncertainty is not. They model aleatoric uncertainty by fitting the variance of Gaussian distributions (also proposed in recent work on lightweight probabilistic extensions for deep networks~\cite{gast2018light}). However, this might lead to unstable training and suboptimal performance. More importantly, this ignored the fact that depth distributions are multi-modal in many cases (for example at depth discontinuities and reflective surfaces). They capture epistemic uncertainty by Bayesian neural networks \cite{gal2016dropout}. However, it requires expensive Monte Carlo sampling to obtain depth predictions and uncertainty estimations. Instead, we focus on modeling the multi-modal distributions over depth, which gives us more reliable uncertainty metrics without the additional computational overhead.

{\bf Multiple Hypotheses Learning (MHL):}
Prior works \cite{guzman2012multiple,lee2016stochastic} formulate the problem of learning to predict a set of plausible hypotheses as multiple-choice learning. They train an ensemble of models to produce multiple possibilities and define an oracle to pick up the best hypothesis. Rupprecht et al.~\cite{rupprecht2017learning} uses a shared architecture to produce multiple hypotheses and train the network by assigning each sample to the closest hypothesis. Different from these approaches, we train a single network to produce a multi-modal distribution, from which we can obtain multiple predictions without directly optimizing an oracle loss in training.

\section{Method}
\label{sec:method}

We solve the problem of inferring continuous depth through discrete classification. To illustrate the method, we first introduce how we discretize continuous depth into discrete categories. Then we show the formulation of depth estimation as a multi-class classification task (mutual exclusive) and a multi-label (binary) classification task (not mutually exclusive). Then we discuss the output of our model, i.e. a probabilistic categorical distribution over discrete depths, and how we will evaluate the output, including evaluating as a standard depth estimation task and as a depth estimation with uncertainty.

{\bf Discretization:}
We discretize continuous depth values in the log space. Given a continuous range of depth $[a,b)$, we discretize it into $K$ intervals, i.e. $[d_{1}, d_{2}), [d_2, d_3), .. , [d_{K}, d_{K+1})$, with
\begin{equation}
d_k = \log a+\frac{k-1}{K}(\log b-\log a), k\in\{1\ldots K\}.
\end{equation}
This captures the perceptual difference in human visual systems, i.e., we care more about differences in depths of close objects than distant ones. Furthermore, due to sensor sampling effects, we tend to encounter more close points rather than far away ones. Working in log space partially alleviates this class imbalance problem.

\begin{figure*}
  \centering
  \includegraphics[trim=0 20 0 0, clip,width=1.0\linewidth]{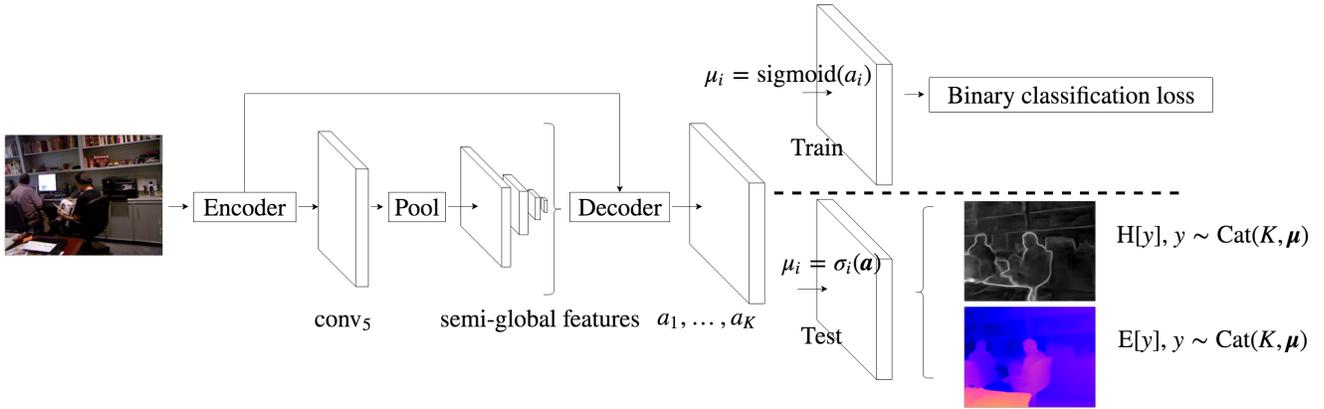}
  \setlength{\belowcaptionskip}{-14pt}
  \caption{Our network architecture consists of an encoder, a spatial pyramid pooling module, and a decoder. Our encoder is a ResNet-50 truncated before global pooling. Spatial pyramid pooling takes ResNet feature then extracts global and semi-global feature through multi-scale pooling. The decoder processes pooled feature to predict a un-normalized score map for each discrete depth class $a_1,\dots,a_K$. During training, each un-normalized score map is pushed into a per-pixel soft-labeled binary cross entropy loss; at test time, we perform per-pixel normalization using softmax across all depth classes to ensure a valid per-pixel distribution over depth, from which we can make the final prediction of depth and uncertainty.  Original-resolution images are used as input, and the predictions are bilinearly up-sampled to the same resolution as ground-truth. }
  \label{fig:arch}
\end{figure*}

{\bf Multi-class Classification:}
As a baseline method, we first show how we recast the continuous regression as a multi-class classification problem. A discrete distribution over depth $y$ can be parameterized by a categorical distribution $Cat(K,{\boldsymbol\mu})$. We learn to predict the probability $\mu_k=p(y=k)$ of each depth label by minimizing the negative log likelihood. Since we use the output of a softmax layer as the predicted probability, we will also refer to this variant as ``Softmax'' in the following text. Given a ground truth label $y^*$, image feature ${\bf x}$, and the model parameters ${\bf w}$, the loss function can be written as,
\begin{equation}
 L({\boldsymbol \mu}|y^*) = -\sum_{k=1}^{K} {\mathds{1}(k=y^*)} \log\mu_k({\bf x};{\bf w}).
 \label{equ:multi}
\end{equation}
Here the distribution ${\boldsymbol\mu}({\bf x};{\bf w})$ is predicted from a $K$-way multi-class classifier.

Equation \eqref{equ:multi} gives us the cross-entropy between an one-hot label vector $\mathds{1}(k=y^*)$ and the predicted distribution ${\boldsymbol\mu}({\bf x};{\bf w})$. To incorporate the ordinal nature of the depth labels, i.e., penalize predictions closer to the ground truth less than predictions further away, we replace the one-hot target vector $\mathds{1}(k=y^*)$ with a discretized Gaussian centered around the ground truth, i.e.,
\begin{align}
q(k;y^*) = \frac{1}{Z}e^{\frac{||k-y^*||^2}{-2\sigma^2}},\end{align}
where $Z$ is the partition function.


{\bf Binary Classification:}
To alleviate competition between depth classes, we further model continuous depth as a collection of $K$ independent Bernoulli random variables $y_k\sim B(1,\mu_k)$, where $\mu_k$ encodes the probability of falling into the $k_{th}$ depth interval. We also refer to this variant as multilabel in the paper. The loss function is written as,
\begin{equation}
\begin{split}
 L({\boldsymbol \mu}|y^*) =& -\sum_{k=1}^{K} [ {\tilde{q}(k;y^*)} \log\mu_k({\bf x};{\bf w}) \\&+ (1-{\tilde{q}(k;y^*)}) (\log (1 - \mu_k({\bf x};{\bf w}))) ],
 \end{split}
 \label{equ:binary}
\end{equation}
where $\tilde{q}(k;y^*) = e^{\frac{||k-y^*||^2}{-2\sigma^2}}$ is an {\it unnormalized} version of soft target distribution. 

One can see this as a {\em relaxation} of the training objective from Eq.\eqref{equ:multi} that drops the constraint that $\sum_k \mu_k = 1$~\cite{li2018brute}. 
The variance $\sigma^2$ is designed such that for all depth classes within $25\%$ difference to ground truth, their label is greater than $0.5$. In test time, we push the pre-logit scores of each binary classifier through a softmax and obtain a distribution over discrete depth, as shown in Fig. \ref{fig:arch}.


{\bf Predicting Depth from a Distribution:}
After obtaining the distribution over depth, Cao et al.~\cite{cao2017estimating} report the most confident depth class, ignoring the multi-modal nature of the predicted distribution.
Different from their approach, we report the {\em expected} depth based on the predicted distribution as $\mathbb{E}[y] = \sum_{k}\mu_kd_k$, which takes into account the whole distribution and yields better depth estimations.

{\bf Uncertainty and Multiple Hypotheses:}
We now describe various statistics that can be computed from our multimodal distribution, motivated by autonomous robotic perception.
Because the perception module of robots needs to be self-aware enough to report potential failures to the downstream planner or online-mapping module when faced with ambiguous scenes, the first statistic is uncertainty, as computed with Shannon entropy:
\begin{equation}
 H(y) = -\sum_k \mu_k\log{\mu_k}.
\end{equation}



Secondly, even if the most-likely (or expected) depth of a particular pixel is far away, a robotic motion planner may wish to decrease speed if there is a non-negligible probability that its depth is in the near-field (due to say, a translucent obstacle). As such, our network can directly output  {\em multiple} depth modes to downstream planners.

{\bf Evaluation:}
Evaluating the above functionality on a robotic platform is difficult. Instead, to evaluate the quality of uncertainty estimation, we make use of the area under ROC curve (AUC), which is widely used in stereo vision and optical flow \cite{bruhn2006confidence,hu2012quantitative}.
To assess the accuracy of the multi-hypotheses output, we follow past work on MHL \cite{guzman2012multiple,lee2016stochastic} and use an ``oracle" evaluation protocol where an algorithm is allowed to report back multiple depth predictions, and the best one is chosen to compute the accuracy \cite{guzman2012multiple}. We also report standard metrics \cite{eigen2014depth} on depth estimation benchmarks. 

\begin{figure}
\centering
\includegraphics[height=2.6cm]{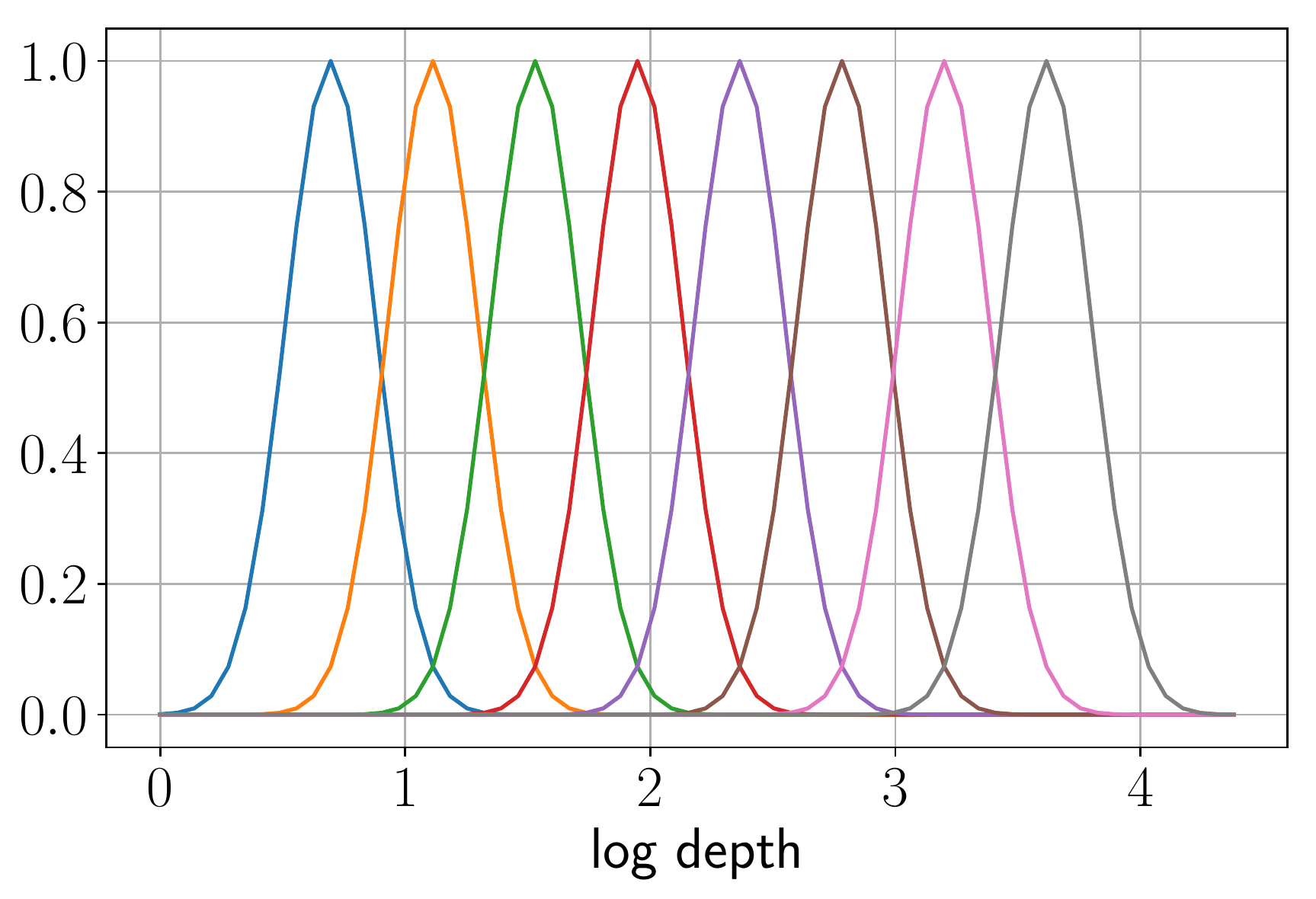}
\includegraphics[height=2.6cm]{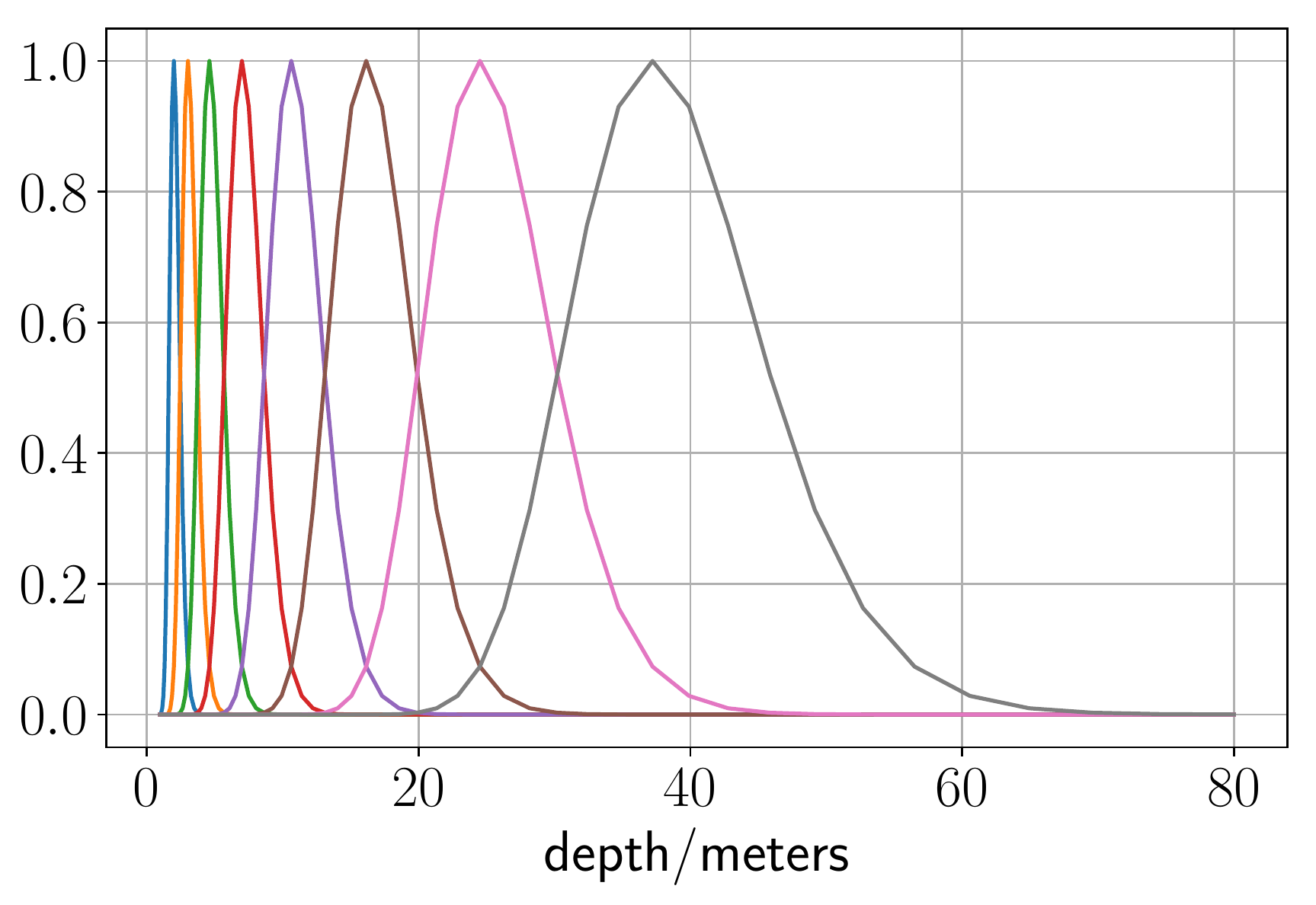}

\caption{Soft target distributions for binary classification in log scale (left) and linear scale (right). We plot the soft target centering on the $\{11,17,23,29,35,41,47,53,59\}$th depth interval.}
    \label{fig:binning}
\end{figure}

{\bf Implementation}
We follow the architecture of Kuznietsov et al.~\cite{kuznietsov2017semi} as shown in Fig.~\ref{fig:arch}. We further add a spatial pyramid pooling module \cite{he2014spatial} to extract global and semi-global features from the scene.
We experimented with different numbers of bins on KITTI. With {32, 64, 96, 128} bins, our method achieves an absolute relative error (ARE) of {9.34\%, 8.61\%, 8.60\%, 8.59\%}. As improvement becomes marginal, we pick 64 as the number of bins and used it for all experiments in this paper. 
Fig. \ref{fig:binning} shows the unnormalized soft-target distribution we use when training binary classifiers.

\section{Experiments}

We first introduce our experimental setup, including dataset and training details. We then compare to prior estimation methods that reason about uncertainties. Finally, we compare our method with the state-of-the-art on the standard depth estimation task, as well as using multi-hypotheses evaluation \cite{rupprecht2017learning}.

{\bf Setup:}
We test our method on the standard depth estimation benchmarks, including KITTI~\cite{geiger2013vision} for outdoor scenes (1-80m) and NYU-v2~\cite{eigen2014depth} for indoor scenes (0.5-10m). On KITTI, we follow Eigen's split~\cite{eigen2014depth} for training and testing. On NYU-v2, we sample $13$k images following~\cite{laina2016deeper} for training and test on the official test split. 


{\bf Training:}
We first initialize the weights of our ResNet-50 backbone with the ImageNet pre-trained ones. To augment training data, we apply random gamma, brightness, and color shift, as in \cite{godard2017unsupervised}. We fine-tune the weights with an Adam optimizer with an initial learning rate of $0.0001$ and decrease the learning rate with a factor of $0.1$ after $45$ epochs. We train our KITTI model for a total of 60 epochs and our NYU-v2 model for a total of 160 epochs. Our experiments are run on a machine with GeForce GTX Titan X GPU using Tensorflow.

\begin{figure}[t]
  \centering
  \includegraphics[width=.48\linewidth]{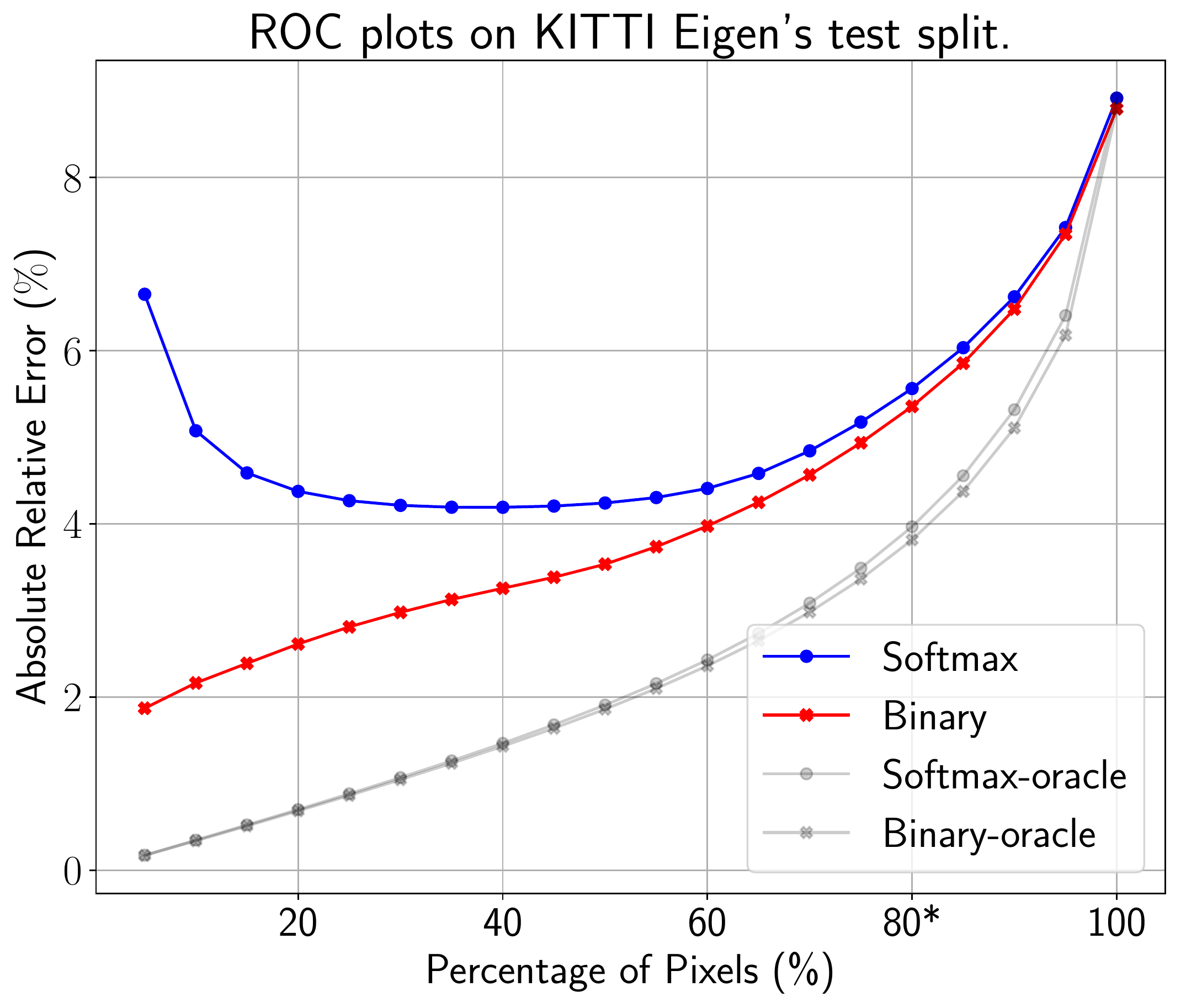}
  ~
  \includegraphics[width=.48\linewidth]{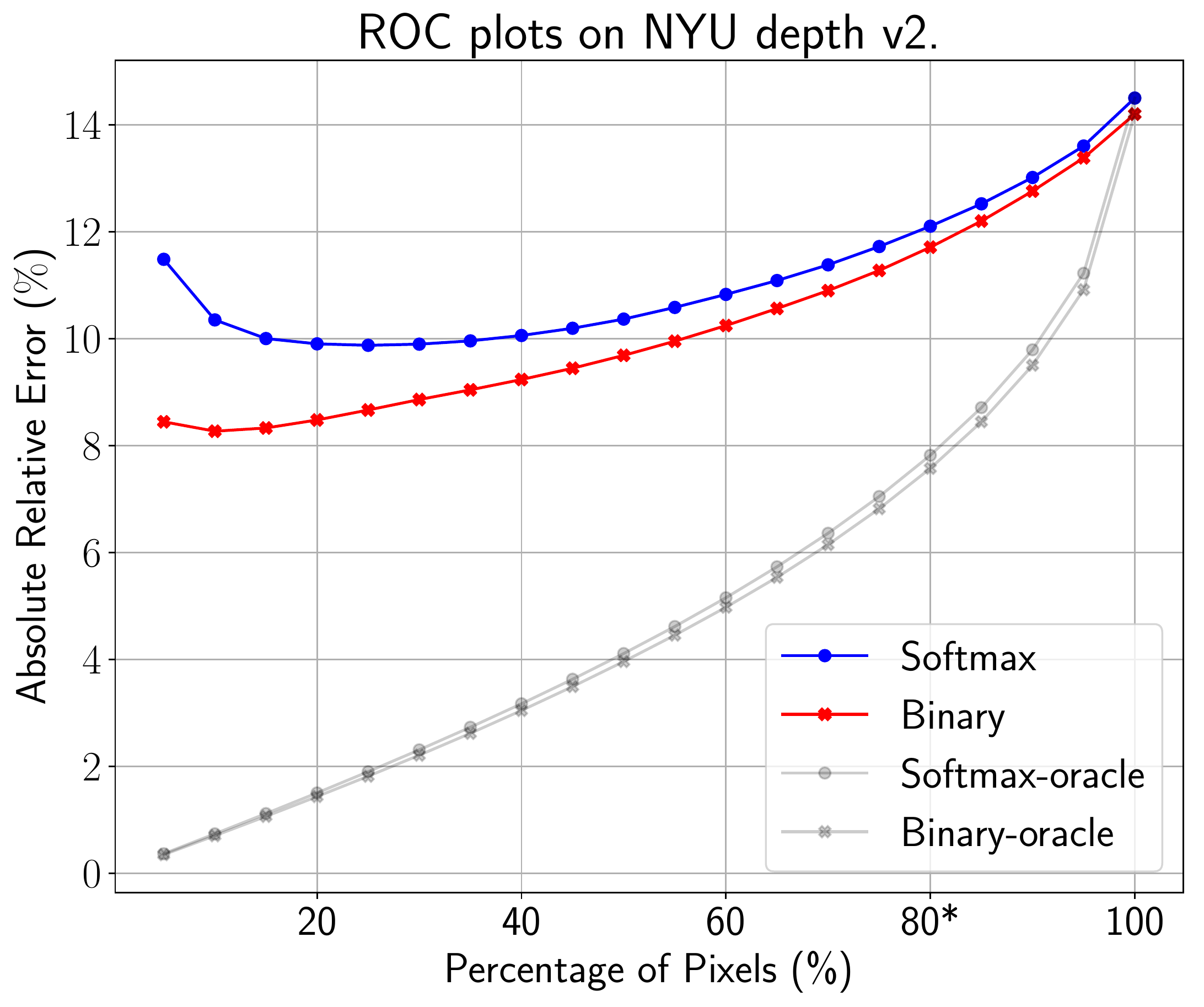}

     \setlength{\belowcaptionskip}{-14pt}
 \caption{How well does the predicted uncertainty correlate with the actual depth estimation performance? 
 We first sort all predictions in ascending order of uncertainty. Then we gradually include more predictions for evaluation by increasing the uncertainty threshold (including more uncertain predictions in the evaluation). The X-axis represents the percentage of pixels we include and Y-axis represents the ARE on the selected pixels.
 Notice uncertainties estimated by the model trained with multi-class classification loss (``Softmax''~\cite{cao2017estimating}) are not well correlated with error, especially for the most confident pixels. On the contrary, the error increases monotonically as confidence drops for our proposed approach (``Binary''). At 80\%, our method also achieves a lower error rate (5.4\% vs. 5.6\%).}
 \label{fig:roc1}
\end{figure}



\subsection{Depth Estimation with Uncertainty}
{\bf Baselines:}
Considering most prior art do not reason about uncertainty, we compare to predictive Gaussian and predictive Gaussian with Monte Carlo dropout (Gaussian-dropout)~\cite{gast2018light,kendall2017uncertainties} in terms of depth estimation with uncertainty, as shown in Tab.~\ref{tab:uncertainty-kitti}. 
For a fair comparison, we re-implement and train predictive Gaussian and Gaussian-dropout on KITTI and NYU depth v2. We make sure the re-implemented version has an architecture that is as close as possible to ours. For predictive Gaussian, we use the same backbone architecture but with a different prediction head, which predicts the mean and variance of a Gaussian distribution over depth in log space. To train predictive Gaussian, we minimize the per-batch negative log-likelihood based on the predicted mean and variance. For Gaussian-dropout, we use the same backbone architecture and prediction head except we perform dropout with a probability of 0.5 after several convolutional layers, as in Kendall et al.~\cite{kendall2015bayesian}. During inference, we draw 32 samples to make predictions and estimate uncertainty. Following the same idea, we apply Monte Carlo dropout to our binary model, referred to as Binary-dropout.

\begin{figure}
  \centering
  \includegraphics[width=.48\linewidth]{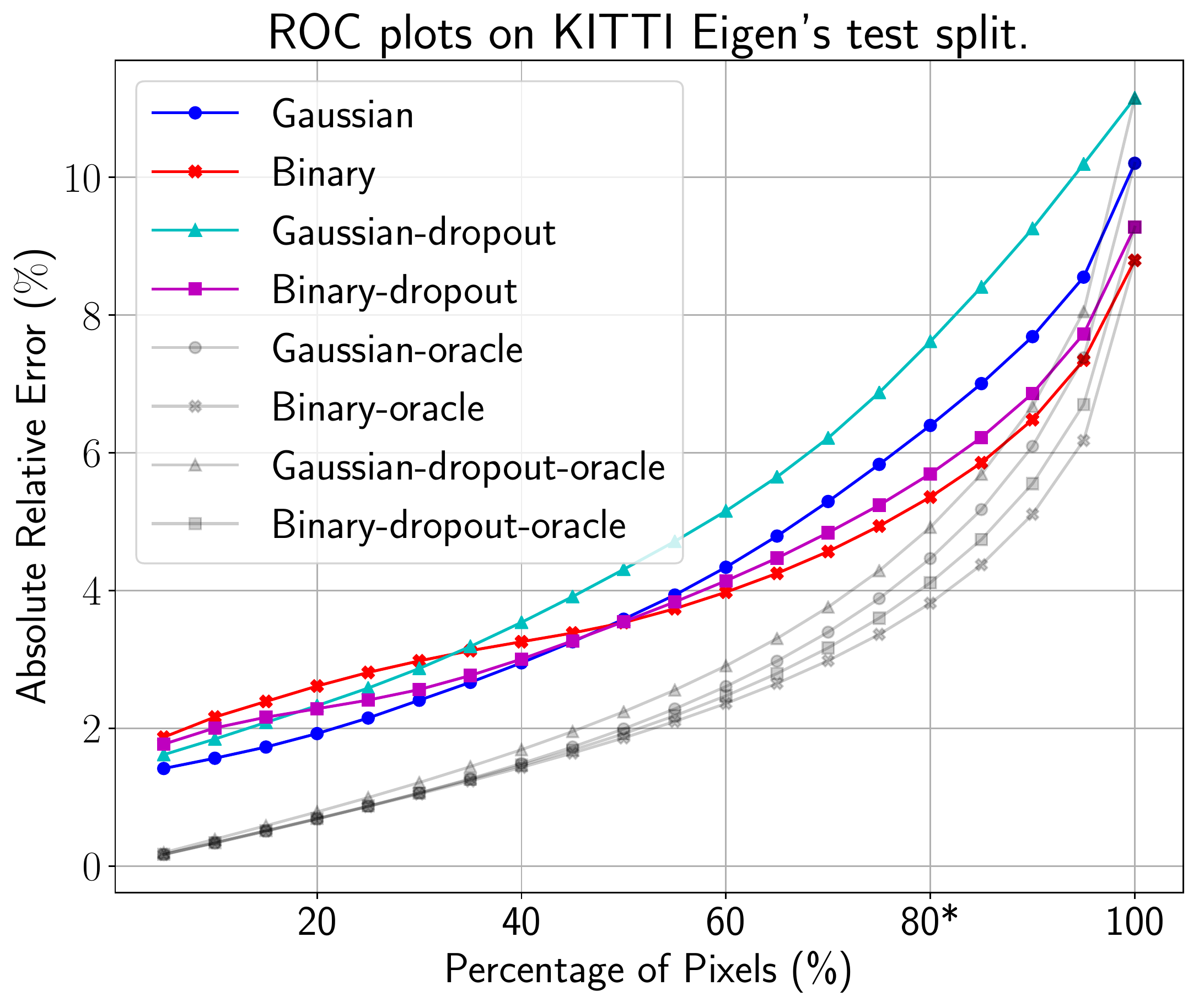}
  \includegraphics[width=.48\linewidth]{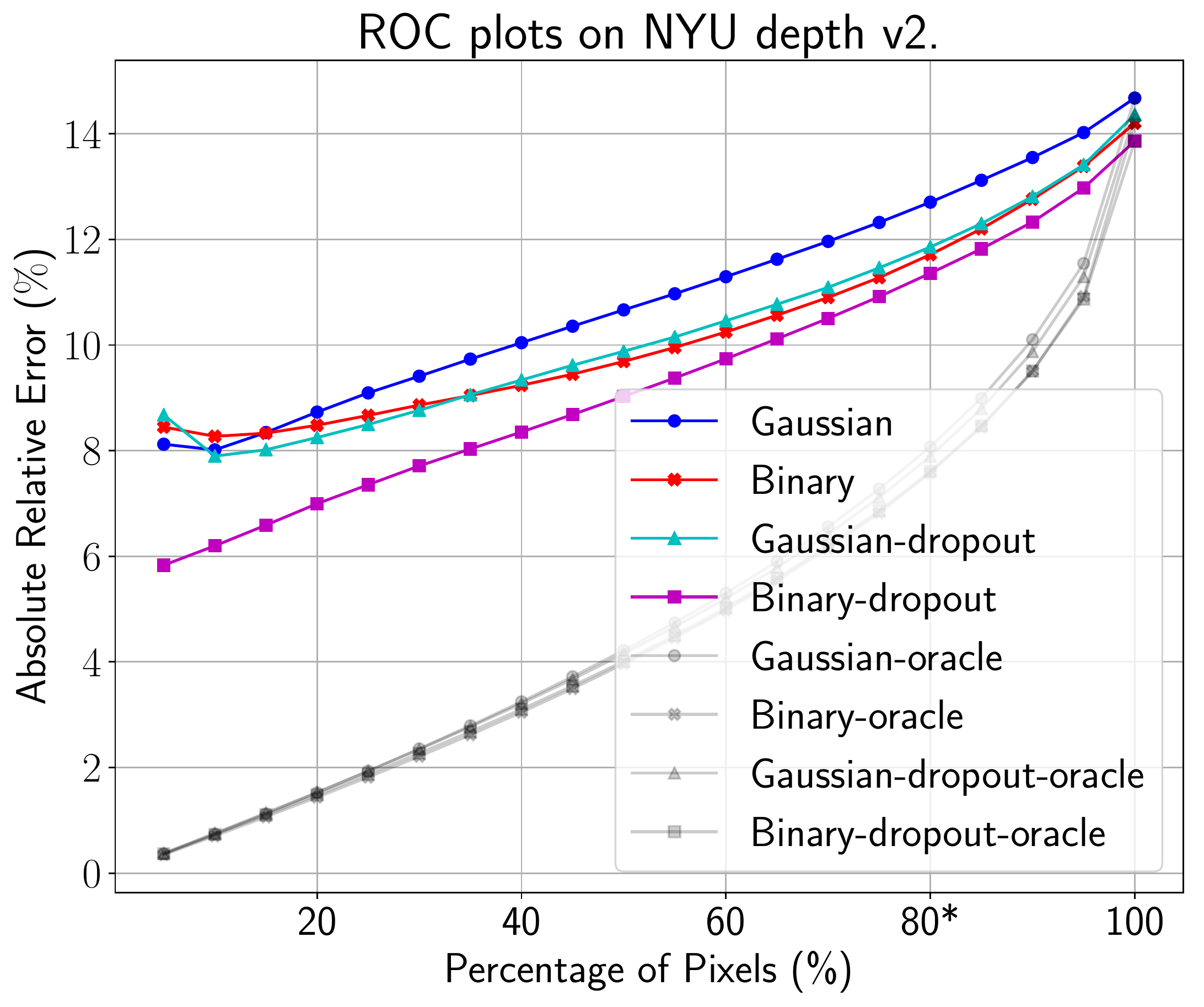}
  \setlength{\belowcaptionskip}{-14pt}
 \caption{
 Compared to predictive Gaussian~\cite{kendall2017uncertainties} (``Gaussian''), our method (``Binary'') yields lower error rate when more than $50\%$ pixels are kept for KITTI, and more than $15\%$ pixels for NYU. By applying Monte Carlo dropout, both predictive Gaussian (``Gaussian-dropout'') and our approach (``Binary-dropout'') see a significant improvement on NYU. While on KITTI, the performance get strictly worse for predictive Gaussian.} 

 \label{fig:roc2}
\end{figure}

Following Hu et al.~\cite{hu2012quantitative}, we plot ROC curves to evaluate our depth estimation with uncertainty, as shown in Fig.~\ref{fig:roc1} and Fig.~\ref{fig:roc2}. Such curves demonstrate how well the predicted uncertainty correlates with the actual depth estimation performance. A point $(x,y)$ on the curve indicates a performance of $y$ on the least uncertain $x$~(\%) predictions over all pixels in the test set. Perfect uncertainty estimation, from the perspective of the ROC curve, should rank predictions as if they are ranked by the actual error. As a reference, we include curves with such oracle w.r.t. a specific error metric (absolute relative error or ARE). Below, we first compare two variants of our model (binary classification and multiclass classification). Then we will compare our model to prior art that predicts uncertainty (predictive Gaussian and Gaussian-dropout). For each sub-metric under AUC, we follow the definition in Eigen et al.~\cite{eigen2014depth}.

{\bf Binary classification vs Multiclass classification:}
In Fig.~\ref{fig:roc1}, we compare the model trained with binary classification loss (``Binary'') to the model trained with multi-class classification loss (``Softmax''). As we can see on the left side of both plots, the uncertainty predicted by the multi-class classifier does not correlate well with the actual error rate, especially for those least uncertain (or most confident) pixels. In contrast, the model trained with binary classification loss produces a curve that monotonically increases as the uncertainty threshold goes up, because it is able to {\em correctly rank more correct pixels as more confident}. We posit that our multilabel loss (that removes a linear constraint present in the multi-class formulation) acts as an additional regularizer that improves uncertainty estimation.


{\bf Gaussian vs Binary:}
In Fig.~\ref{fig:roc2}, we find predictive Gaussian also yields reliable uncertainty estimation, as it produces a monotonically increasing curve. Overall it achieves a slightly worse performance, comparing to our model trained with binary classification. It might be due to its uni-modal assumption and optimization difficulties in training time (discussed further in our ablation study). Interestingly, adding Monte Carlo dropout significantly improves NYU performance for both predictive Gaussian (``Gaussian-dropout'') and our approach (``Binary-dropout''). However, on KITTI, we see a strictly worse performance for the predictive Gaussian.

{\bf Quantitative evaluation:}
In Tab.~\ref{tab:uncertainty-kitti}, we further compare uncertainty estimation quantitatively using metrics introduced in Section~\ref{sec:method}. Our binary classification method produces better performance in terms of AUC compared to predictive Gaussian and its Monte Carlo dropout variant in terms of ARE and $\delta_1$, without expensive Monte Carlo sampling. By adding Monte Carlo dropout to our model, we can further improve AUC of ARE, RMSE and $\delta_1$ on NYU depth v2. Although predictive Gaussian with Monte Carlo dropout outperforms our binary loss on all metrics based on RMSE, it is too slow for real-time perception. 
 Please refer to Tab.~\ref{tab:uncertainty-kitti} for more detailed discussion.

\begin{table}[t]
\centering
\begin{tabular}{cccccc}
\toprule
 &&\multicolumn{3}{c}{AUC} & time\\
\cmidrule(lr){3-5}
& Method & ARE & RMSE & $1-\delta_1$ & (ms)\\
\midrule
\multirow{5}{*}{K}
&Gaussian~\cite{kendall2017uncertainties}
&4.38 &1.42 &2.63  &64\\
&Softmax
&5.19 &2.88 &2.93  &74\\
&Binary
&\underline{\bf{4.17}} &\bf{1.33} &\underline{\bf{1.79}}  &74 \\
\cmidrule(lr){2-6}
&Gaussian-dropout~\cite{kendall2017uncertainties}
&5.18& \underline{1.21} &3.61  &467\\
&Binary-dropout
&4.20 &1.33 &2.06  &540\\
\midrule
\multirow{5}{*}{N}
&Gaussian~\cite{kendall2017uncertainties}
&10.94 &\bf{0.41} & 10.95 & 44 \\
&Softmax
&11.17 &0.53   &11.09  &52\\
&Binary
&\bf{10.28} &0.42 &\bf{9.26} &52\\
\cmidrule(lr){2-6}
&Gaussian-dropout~\cite{kendall2017uncertainties}
&10.33 &\underline{0.32} &10.30 &  353 \\
&Binary-dropout
&\underline{9.39} &0.40   &\underline{7.79} & 410\\
\bottomrule
\end{tabular}
\setlength{\belowcaptionskip}{-14pt}
\caption{Quantitative evaluation for uncertainty estimation on KITTI~(K) and NYU-v2~(N). The best results among methods without Monte Carlo dropout are made bold, while the best considering Monte Carlo dropout are underlined. On both datasets, we compare our method trained with the binary loss (``Binary'') and the multiclass loss (``Softmax'') to predictive Gaussian~\cite{kendall2017uncertainties} (``Gaussian''). The quantitative results are consistent with Fig.~\ref{fig:roc1} and Fig.~\ref{fig:roc2}. In terms of AUC on ARE and $1-\delta_1$~({\it the lower the better}), our binary loss consistently outperforms predictive Gaussian on both KITTI and NYU-v2. Importantly, when combined with Monte Carlo dropout, our binary model (``Binary-dropout'') further reduces the AUC on NYUv2.}
\label{tab:uncertainty-kitti}%
\end{table}

\begin{table}
\centering
    \begin{tabular}{cccccc}
\toprule
&Method &ARE ($\%$)  &RMSE   &$\delta_1$ ($\%$)& time (ms)\\
\midrule
\multirow{7}{*}{K}
&Binary
&{\bf 8.9} &{\bf 3.85}  &90.7  &74\\
&Fu et al.~\cite{fu2018deep}
&9.1   &3.90  &90.5 &74\\
&Cao et al.~\cite{cao2017estimating}
&9.3  &4.02 & {\bf 90.8} &74\\
\cmidrule(lr){2-6}
&Eigen et al.~\cite{eigen2014depth}
&19.0&7.16  &69.2  &13\\
&Godard et al.~\cite{godard2017unsupervised}
&11.4  &4.94  &86.1  &35\\
&Cao et al.~\cite{cao2017estimating}
&11.5  &4.71  &88.7 &-\\
&Fu et al.~\cite{fu2018deep}
&\underline{7.2}    &\underline{2.73} &\underline{93.2} &1250\\
\midrule
\multirow{7}{*}{N}
&Binary
&14.2  &0.51  &82.7  & 52\\
&Binary-dropout
&{\bf 13.9}  &\underline{\bf 0.50}  &\underline{\bf 82.8}  & 410\\
&Kendall et al.~\cite{kendall2017uncertainties}
&14.4   &0.51  &81.5  & 353\\
\cmidrule(lr){2-6}
&Eigen et al.~\cite{eigen2014depth}
&15.8 &0.64 &76.9  & 10\\
&Laina et al.~\cite{laina2016deeper}
&12.7  &0.57 &81.1 & 55\\
&Fu et al.~\cite{fu2018deep}
&11.5  &0.51  & \underline{82.8} &-\\
&Kendall et al.~\cite{kendall2017uncertainties}
&\underline{11.0}    &0.51  &81.7 & 7500\\
\bottomrule
\end{tabular}
  \caption{Performance on KITTI~(K) Eigen's split and NYU-V2 depth~(N) dataset.
  The best results over the light-weight setup are bolded, while the best results overall are underlined. On KITTI, our method outperforms the state-of-the-art Fu et al.~\cite{fu2018deep} under the same setup. With its original setup (a heavy-weight backbone and test-time ensemble), \cite{fu2018deep} runs nearly 17x times slower (1250ms vs 75ms). On NYU-v2, our method outperforms Kendall et al.~\cite{kendall2017uncertainties} with the same backbone network. With its original setup, Kendall et al.~\cite{kendall2017uncertainties} runs 144x slower. Our method further improves when training with dropout and testing with MC sampling~\cite{kendall2015bayesian}, referred to as Binary-dropout.
}
\label{tab:main-kitti}%
\end{table}

\subsection{Multi-hypothesis Depth Prediction}
We first evaluate standard depth prediction performance on KITTI and NYU-v2 using metrics proposed in~\cite{eigen2014depth}, as shown in Tab.~\ref{tab:main-kitti}. We then extend the evaluation by allowing multiple depth hypotheses. For a fair comparison, we re-implement Fu et al.~\cite{fu2018deep} and Cao et al.~\cite{cao2017estimating} under the same setup as ours (a light-weight backbone and no test-time ensemble). We also include numbers in the original paper as a reference. Please refer to Tab.~\ref{tab:main-kitti} for detailed comparison.

\begin{figure}
  \centering
  \includegraphics[width=.48\linewidth]{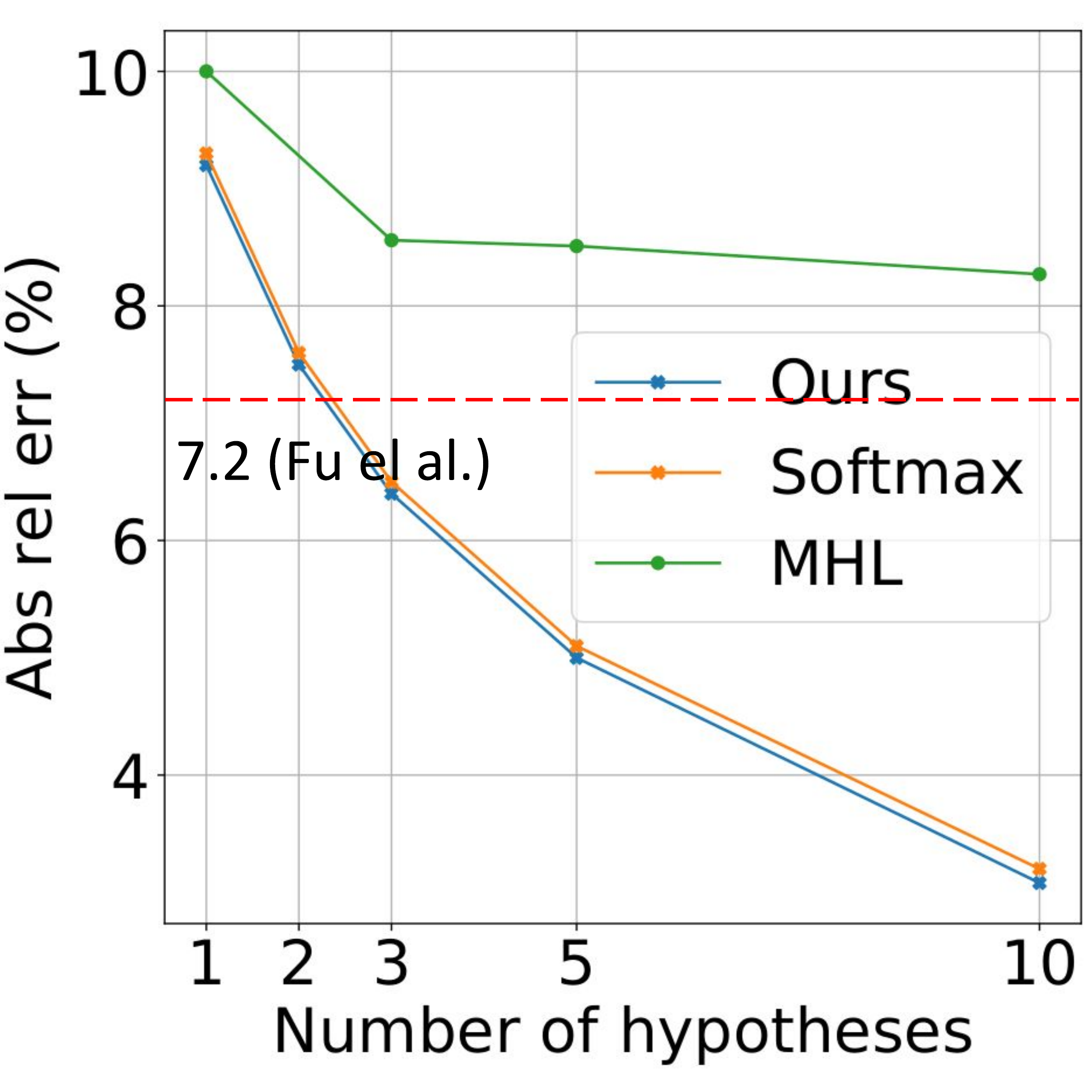}
  ~
  \includegraphics[width=.48\linewidth]{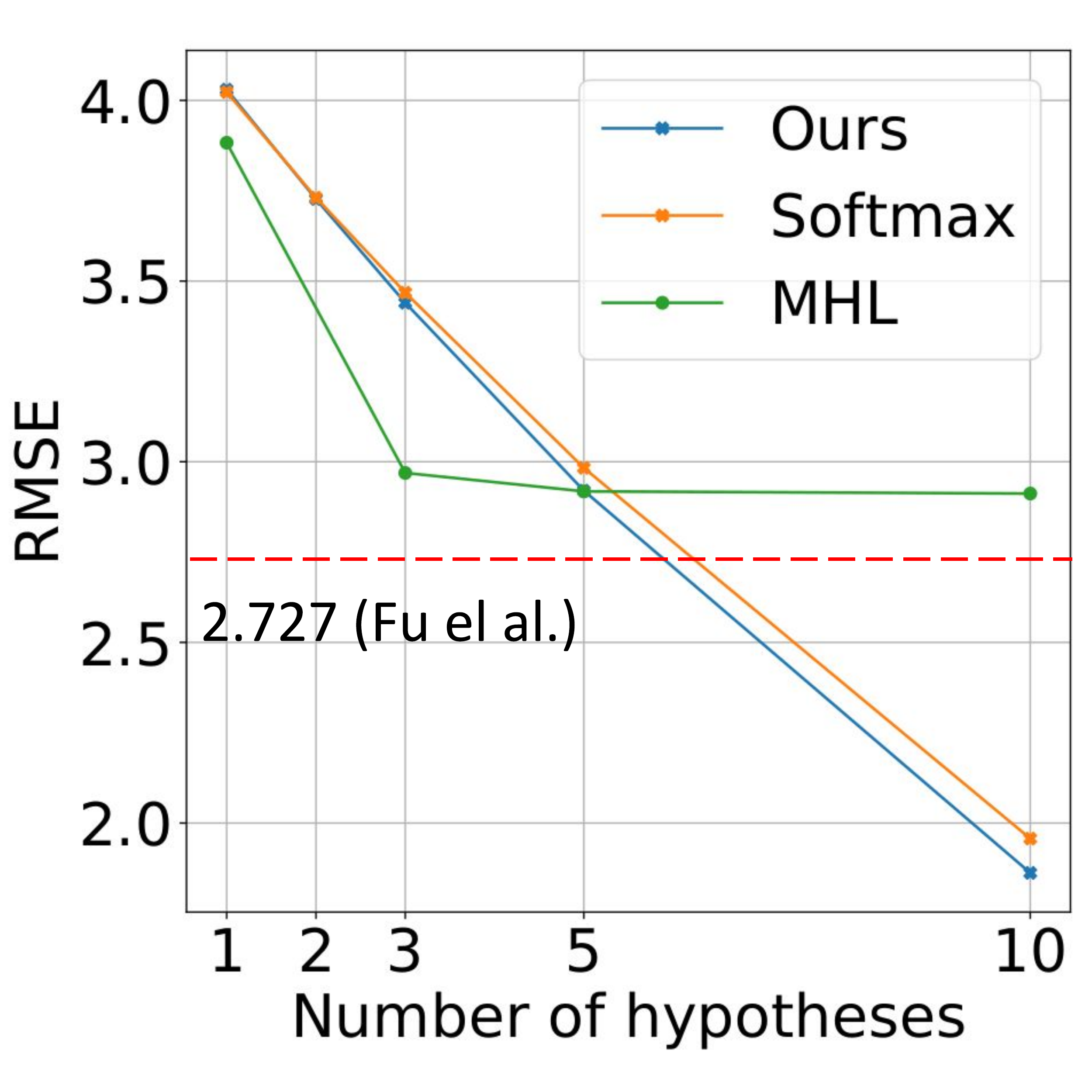}
  
    \setlength{\belowcaptionskip}{-14pt}
  \caption{Error as a function of hypotheses number on KITTI. Compared to MHL, our method always produces better results in terms of ARE. As for RMSE, our method performs worse than MHL when $M<5$, possibly because the MHL baseline is trained to directly minimize squared error. However, MHL's error stops going down after $M\geq 3$, while we do not observe this effect for our model. Compared to softmax (Cao et al.~\cite{cao2017estimating}), our method also achieves slightly better performance. Also, our method consistently out-performs Fu et al.~\cite{fu2018deep} in ARE using more than two hypotheses, and in terms of RMSE using more than five hypotheses.}
\label{fig:kittimhl}
\end{figure}

To evaluate our multi-modal distributions, we follow the standard protocol in multi-hypothesis learning~\cite{lee2016stochastic}. After computing the pre-logits scores, we report back $M$ depth hypotheses with the highest scores, and the one with the lowest error is selected by the oracle for evaluation.

Since most methods {\it can't} output multiple hypotheses, we compare to the ones that {\it can} be trained to output multiple hypotheses~\cite{rupprecht2017learning}, referred to as MHL. Similar to traditional $L_2$ regression, we directly regress to the depth in log space. However in training time, we make $M$ predictions and construct an oracle loss by selecting the prediction that best describes the ground-truth in terms of $L_2$ distance. We train the MHL baseline for $M=1,3,5,10$, and use an oracle to select the best prediction for evaluation. Please see Fig. \ref{fig:kittimhl} for analysis of the results.

\section{Building Maps with Uncertainty}
In this section, we demonstrate one application of geometric uncertainty estimation: robust map reconstruction. Though maps are often constructed in an offline stage, online mapping can be an integral part of autonomous navigation in unknown/changing environments~\cite{ort2018autonomous}.

In practice, is it notoriously difficult to build 3D maps from raw depth predictions because they tend to contain ``streak-like artifacts''~\cite{barsan2018robust}, which not only affect the quality of the map but also increase the memory usage (because they often result in larger occupied volumes). Empirically, we find that such artifacts often happen where ground truth depth is inherently ambiguous and follows a multi-modal distribution, e.g. depth discontinuities and reflective surfaces. Since our depth estimator is designed to predict multi-modal distributions over depth, we use it to improve the accuracy of map reconstruction. By simply thresholding the uncertainty of each pixel's predicted distributions, we can significantly reduce streak artifacts and memory usage, as shown in Fig.~\ref{fig:mapping}. 

We evaluate the performance of map reconstruction with and without uncertainty on KITTI odometry sequence-00~\cite{Geiger2012CVPR}, which is not included in the training set. Specifically, we run our monocular depth estimator on left RGB images, and feed the output depth maps together with ground-truth odometry as the input of Octomap~\cite{hornung2013octomap}. The accuracy is measured as the percentage of correctly mapped map cells, where a cell counts as correctly mapped if it has the same state (free or occupied) as the LiDAR map (ground-truth).  As shown in Tab.~\ref{tab:map-acc}, applying a simple uncertainty-based ranking and selection improves the accuracy of monocular maps by 1.8\% and reduces the memory usage by 25\%.

\begin{table}
\centering
    \begin{tabular}{cccccc}
\toprule
Method &Accuracy (\%) & Memory (MB)\\
\midrule
LiDAR-FOV$^{\dagger}$		&95.9	& 1220.9\\
Ours-binary 	&88.3	& 1682.6\\
Ours-binary-80\%&{\bf 89.9}	& {\bf 1263.2}\\
\bottomrule
\end{tabular}
\setlength{\belowcaptionskip}{-14pt}
  \caption{Accuracy and memory usage of online mapping. LiDAR-FOV indicates the map built using LiDAR points in the left camera field of view, which is the upper-bound of our methods. The map built with top 80\% most confident estimations of our model (Ours-binary-80\%) significantly reduces the memory usage and also improves the mapping accuracy.}
\label{tab:map-acc}%
\end{table}

\section*{Conclusion}
Robotic applications of perception present new challenges for safety-critical, fault-tolerant operation. Inspired by past approaches that advocate a probabilistic Bayesian perspective, we demonstrate a simple but effective strategy of discretization (with the appropriate quantization, smoothing, and training scheme) as a mechanism for generating detailed predictions that support such safety-critical operations.

\appendix
\section{Supplementary material}
\subsection{Ablation study}
To reveal the contribution of each design choice to the accuracy of the standard depth estimation task, we perform an extensive ablation study as shown in Tab. \ref{tab:aba-append}.

{\bf Classification vs	 Regression:}
We first compare $L_2$ regression loss to classification losses (Binary and Multiclass). We find that classification loss always outperforms $L_2$ regression method in terms of absolute relative error and $\delta_1$. However, $L_2$ regression achieves competitive RMSE, likely because it directly minimizes squared error. We also implement Berhu~\cite{laina2016deeper} regression loss, and it is still easily out-performed by classification-based methods.

{\bf Multiclass vs Binary classification}
Training with binary classification loss gets similar performance compared to multiclass classification loss on KITTI. However, it yields significantly better results on NYU. Since test images in NYU differ more from the training images than KITTI, we posit that binary classification loss gives better generalization ability compared to multiclass classification loss.

{\bf Effect of Monte Carlo dropout}
On KITTI, Monte Carlo dropout makes prediction performance worse for both binary classification method and predictive Gaussian. However on NYU, it improves results for both methods. This is possible because NYU contains more diverse scenes, where dropout helps prevent overfitting. While on KITTI, training and testing data are highly correlated. Therefore, regularizing the model by dropout does not help.

{\bf Expectation vs Most-likely class inference} On KITTI, we find that expectation yields better results for all metrics except for $\delta_1$. While for NYU, expectation always out-performs (or on par with) most-likely class. This indicates that expectation is a better way of making a prediction from a depth distribution, since it makes use of the whole distribution.

{\bf Soft targets vs One-hot targets}
Comparing the results of training with soft-target distribution vs one-hot label, we find that soft-target always performs better. We posit that by training with soft targets, our model benefits from sample sharing, and thus performs better than using one-hot labels.

\begin{table}[!h]
  \caption{Results of using a mixed KITTI and NYU-v2 dataset for training. The model is trained with binary classification loss and predicts the most-likely class at test time.}
  \centering
    \begin{tabular}{ccccc}
\toprule
Test dataset &Abs Rel ($\%$)  &RMSE   &$\delta_1$ ($\%$)\\
\midrule
 KITTI 
&9.9  & 3.969 &89.1 \\
 NYU-v2
&15.3  &0.541 &80.3 \\
\bottomrule
\end{tabular}
\label{tab:mix-dataset}%
\end{table}

\subsection{Training on mixed KITTI and NYU-v2}
To obtain a robust model that works for both indoor and outdoor scenes, we train a single model using KITTI and NYU-v2. To precisely capture the full depth range in both datasets, we adjust the depth range to $0.5$m to $80$m and the number of depth intervals to $100$. At training time, we randomly crop the data to $384\times640$ and average loss over the image before averaging over the whole batch. As shown in Tab.~\ref{tab:mix-dataset}, when trained jointly, the performance of our model is not severely affected on both datasets.

{\bf Acknowledgements} This work was supported by the CMU Argo AI Center for Autonomous Vehicle Research.


\bibliographystyle{ieee}
\bibliography{ref}

\begin{table*}[!t]
  \caption{Ablation study where best results are bolded. 
  $^*$As for log scale metrics, we use $\text{RMSE}_{\log}$ for KITTI, and $\log_{10}$ for NYU-v2.}
\makebox[\textwidth][c]{
    \begin{tabular}{crcccc}
\toprule
Dataset &Method &Abs Rel ($\%$) &$\log^*$   &RMSE   &$\delta_1$ ($\%$)\\
\midrule
\multirow{12}{*}{KITTI}&
binary (expectation)
&\bf{8.9} &0.157  &3.847  &90.7\\
&binary-dropout (expectation)
&9.4 &0.159  &3.920  &89.4\\
&binary (most-likely)
&9.2 &0.167  &4.030  &\bf{90.9}\\
&binary-hard label (expectation)
&9.0 &0.160  &3.880  &89.9\\
&binary-hard label (most-likely)
&9.3 &0.174  &4.204  &90.1\\
&multiclass (expectation)
&9.0 &\bf{0.156}  &\bf{3.842}  &90.3\\
&multiclass (most-likely) \cite{cao2017estimating}
&9.3 &0.167 &4.024  &90.8\\
&ordinal \cite{fu2018deep}
&9.1 &0.161  &3.895  &90.5 \\
\cmidrule(lr){2-6}
&$L_2$
&10.0 &0.160  &3.883  &88.9\\
&Berhu~\cite{laina2016deeper}
&10.0  &0.170 &4.113  &89.3\\
&Gaussian~\cite{kendall2017uncertainties}
&10.3 &0.168  &4.11  &87.6\\
&Gaussian-dropout~\cite{kendall2017uncertainties}
&11.3 &0.178  &4.28  &86.2\\
\midrule
\multirow{8}{*}{NYU-v2}
&binary (expectation)
&14.2  &0.059 &0.512 &82.7 \\
&binary-dropout (expectation)
&\bf{13.9} & \bf{0.058} & \bf{0.499} & \bf{82.8}\\
&binary (most-likely)
&14.5 &0.059 & 0.527 &82.4 \\
&multiclass (expectation)
&14.5  &0.060 &0.523 &81.9 \\
&multiclass (most-likely)~\cite{cao2017estimating}
&14.6 & 0.061 &0.543 & 81.7 \\
\cmidrule(lr){2-6}
&Gaussian~\cite{kendall2017uncertainties}
&14.7 & 0.061 & 0.517 & 81.0\\
&Gaussian-dropout~\cite{kendall2017uncertainties}
&14.4 & 0.060 & 0.509 & 81.5\\
\bottomrule
\end{tabular}
}
\label{tab:aba-append}%
\end{table*}

\end{document}